\documentclass{vgtc}                          

\graphicspath{{figures/}{pictures/}{images/}{./}} 

\usepackage{times}                     

\usepackage{tabu}                      
\usepackage{booktabs}                  
\usepackage{lipsum}                    
\usepackage{mwe}                       

\usepackage{mathptmx}                  

\onlineid{0}

\vgtccategory{Research}

\vgtcinsertpkg

\title{Aiding Humans in Financial Fraud Decision Making:\\ Toward an XAI-Visualization Framework} 

\author{Angelos Chatzimparmpas\thanks{e-mail: a.chatzimparmpas@uu.nl}\\ %
        \scriptsize Utrecht University, NL %
\and Evanthia Dimara\thanks{e-mail: evanthia.dimara@gmail.com}\\ %
     {\scriptsize \centering Utrecht University, NL}}

\teaser{
  \centering
  \vspace{-0.9em} \includegraphics[width=0.99\linewidth]{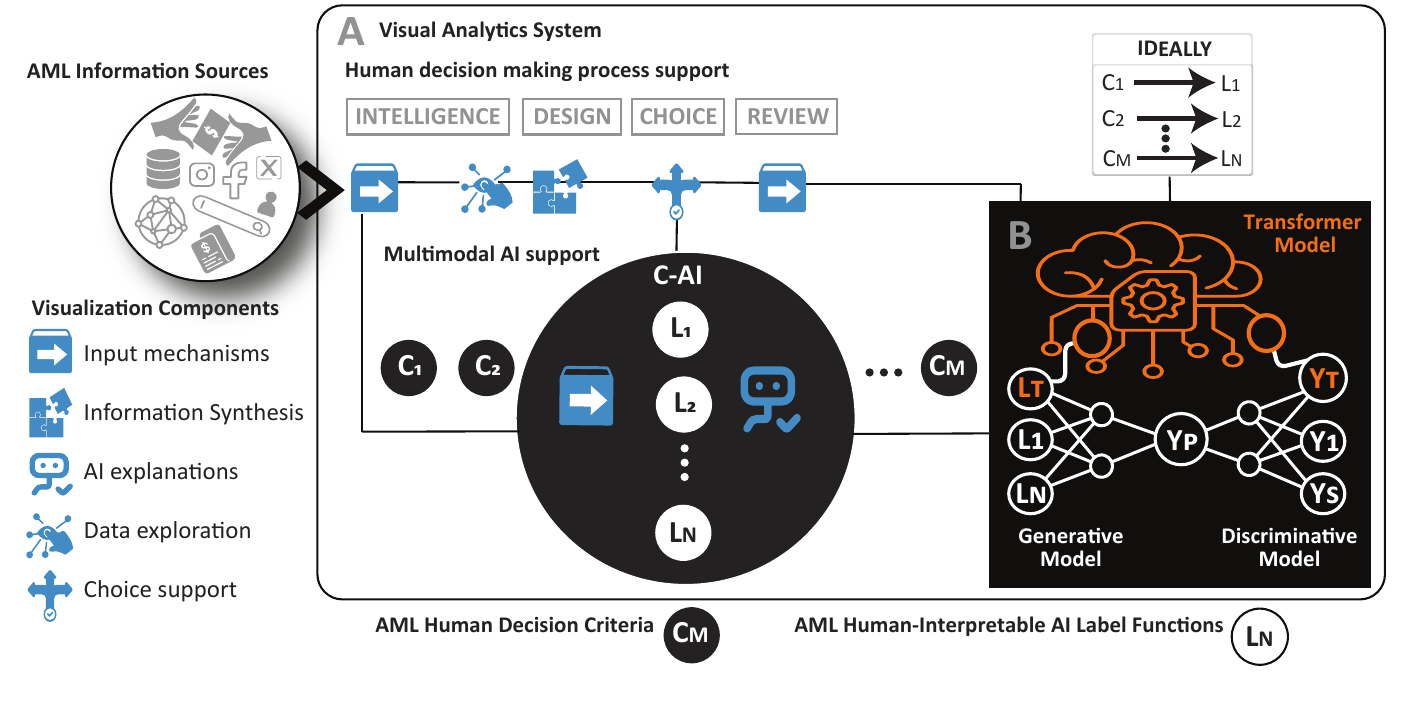} \vspace{-8mm}
   \caption{
The proposed framework features a Visual Analytics (VA) system to support the Anti-Money Laundering (AML) decision making process. The VA system implements various input mechanisms to gather and synthesize information from diverse data sources and employs eXplainable AI (XAI) methods to provide transparent AI support. By leveraging weakly supervised learning, the system combines human expertise with AI to establish decision making guidelines, such as explaining why certain client transactions are flagged as suspicious and how to learn from these fraudulent cases to refine the apriori AML strategies.
}
  \label{fig:teaser} 
}

\abstract{

AI prevails in financial fraud detection and decision making. Yet, due to concerns about biased automated decision making or profiling, regulations mandate that final decisions are made by humans. Financial fraud investigators face the challenge of manually synthesizing vast amounts of unstructured information, including AI alerts, transaction histories, social media insights, and governmental laws. Current Visual Analytics (VA) systems primarily support isolated aspects of this process, such as explaining binary AI alerts and visualizing transaction patterns, thus adding yet another layer of information to the overall complexity. In this work, we propose a framework where the VA system supports decision makers throughout all stages of financial fraud investigation, including data collection, information synthesis, and human criteria iteration. We illustrate how VA can claim a central role in AI-aided decision making, ensuring that human judgment remains in control while minimizing potential biases and labor-intensive tasks.

} 


\begin{document}



\firstsection{Introduction}
\maketitle

The United Nations Office on Drugs and Crime estimates that 2-5\% of the global GDP -- equating approximately \$800 billion to \$2 trillion -- is laundered annually \cite{Europol2024MoneyLaundering}, leading to the development of numerous Anti-Money Laundering (AML) systems that use Artificial Intelligence (AI) \cite{al-suwaidi_anti-money_2020}. Yet, despite the unprecedented data complexity and volume, it remains a legal requirement that human agents make the final AML decisions \cite{Han_et_al_2}, ensuring compliance and accountability \cite{EU2024AIact}, within the frequent and worrisome fraudulent AI outcomes \cite{Abdulalem2022MLFinancialFraudReview}. Although the need for accurate human judgment over complex AML data presents a great opportunity for Visual Analytics (VA), current VA systems only support small and isolated parts of the decision process, either explaining AI predictions like in \cite{Collaris2018Instance} or complementing AI with visual pattern detection (e.g., \cite{lin_taxthemis,zhou_fraud}).

We identify several reasons that prevent visualization solutions from playing a leading role in AML. First, AML decision makers are expected to collect and synthesize numerous heterogeneous and often unstructured data sources, including legal documents, regulatory reports, sanctions lists, and harnessing social media and web content \cite{Han_et_al_2}, while visualization researchers have only recently begun to reflect on the lack of such input design mechanisms \cite{Bressa2024Input}. Second, AML is essentially a decision making task \cite{Gao_Xu}, which recent reviews reveal has not yet been well-understood in visualization research, neither conceptually \cite{Dimara_Stasko} nor empirically \cite{Oral2024Decoupling}, while only a handful of decision-focused visualizations have been developed in academic works \cite{Oral2024Choice} and used in real-world contexts \cite{Dimara2021Unmet}. Third, despite the valuable VA research on eXplainable AI (XAI), AML decision makers find it challenging to adopt complex systems that rely on post hoc XAI methods \cite{Chatzimparmpas2024Visualization}. Finally, the dynamic nature of AML problems often misaligns with the types of AI used -- e.g., (semi-) supervised learning that may involve cumbersome and time-consuming data labeling processes -- thereby hindering XAI-visualization research from providing effective solutions \cite{Chen_et_al.}.

\section{Developing an XAI-VIS Framework for AML}

To address the aforementioned issues, in this poster abstract, we propose a multi-aspect framework that extends Gao and Xu's \cite{Gao_Xu} AML specification of Simon's decision making conceptual model \cite{Simon}. \autoref{fig:teaser} illustrates that extension, indicating our XAI approach for decision support (black circle and related black box) as well as a four-stage (i.e., intelligence, design, choice, and review) contribution of visualization aids (shown in blue).

\vspace{0.4em}
\noindent \texttt{XAI approach}: AML decision making is fragmented and inefficient, as AI primarily analyzes only a subset of data sources, such as client transaction data, while decision makers still have to manually analyze additional data sources to establish decision criteria ($C_{1}, C_{2}, ..., C_{N}$) \cite{Gao_Xu}. This subset analysis by AI, which outputs only binary fraudulent or non-fraudulent results, often remains unclear, leading to unnecessary or insufficient AML investigations and not aligning with the comprehensive AML criteria of decision makers.
To the best of our knowledge, our framework is the first to feature a multimodal, weakly supervised AI \cite{Ratner2017Snorkel} which oversees the four decision making stages and facilitates user interaction through \textit{white box label functions} (e.g., rules, keywords, heuristics). The XAI method \cite{Lang2021Self} consists of three deep learning models: generative, discriminative, and transformer (see \autoref{fig:teaser}B). The \textit{generative model} combines the human-interpretable AI label functions ($L_{1}, L_{2}, ..., L_{N}$) to probabilistically label client actions as fraudulent or non-fraudulent ($Y_{P}$). The \textit{discriminative model} uses a noise-aware loss function to generalize from the data subsample with probabilistic labels ($Y_{P}$) to the entire sample population ($Y_{1}, Y_{2}, ..., Y_{S}$). The \textit{transformer model} serves as a pre-trained, automatic agent that suggests new label functions (e.g., $L_{T}$) and highlights conflicting ones from $L_{1}, L_{2}, ..., L_{N}$. The loop closes with predictions from both the transformer and discriminative models for all samples (noted as $Y_{T}$ \textit{vs} $Y_{1}$ for a single client shown in \autoref{fig:teaser}B) and with further fine-tuning and AI-aided conflict resolution. Ultimately, instead of AI being interpreted as a single decision criterion, the decision criteria ($C_{1}, C_{2}, ..., C_{N}$) set by a human decision maker (e.g., rules based on governmental laws) align perfectly with the label functions ($L_{1}, L_{2}, ..., L_{N}$), as shown in \autoref{fig:teaser}A, top right. As a result, the iterative process enables our black box AI models to facilitate not only efficient but also transparent decision making.

\vspace{0.4em}
\noindent \texttt{Visualization aids per stage}: VA systems incorporating our XAI approach can assist users throughout the four stages of the decision making process (\autoref{fig:teaser}A). The \textsc{intelligence} stage \cite{Gao_Xu} is implemented via \textit{input mechanisms} that transform prior knowledge and raw data sources -- whether existing or mined -- into structured data, either automatically or manually aided by input visualization design \cite{Bressa2024Input,Collins2022Visual}. Users can then \textit{explore the data}, leading to the \textsc{design} stage \cite{Gao_Xu}, where the VA system enables \textit{information synthesis} through structured data verification and multimodal AI-supported inspection, as well as simulations of alternative what-if scenarios. The VA system serves as a mediator between the AI black box and the human decision criteria ($C_{1}, C_{2}, ..., C_{N}$). During the \textsc{choice} stage \cite{Gao_Xu}, the \textit{AI explanations} are in the form of human-interpretable AI label functions for effective \textit{choice support}. During the \textsc{review} stage \cite{Gao_Xu}, decision makers can also reflect on past choices, and the reports of convicted clients can be fed back to the multimodal AI support as ground truth (GT) labels via the extra \textit{input mechanism}, leading to further improvements of the label functions, GT labels, and tuning of the AI models involved. Finally, decision makers can gain new knowledge from the transformer model's label functions ($\Sigma$$L_{T}$) and incorporate their external knowledge by defining new label functions ($L_{1}', L_{2}', ..., L_{N}'$) via novel input visualizations (e.g., as described in \cite{Bressa2024Input,Collins2022Visual}).

\bibliographystyle{abbrv-doi-narrow}

\bibliography{template}
\end{document}